\pdfoutput=1
\documentclass[11pt]{article}

\usepackage[final]{acl}

\usepackage{times}
\usepackage{latexsym}

\usepackage[T1]{fontenc}
\usepackage[utf8]{inputenc}

\usepackage{microtype}

\usepackage{inconsolata}

\usepackage{graphicx}

\usepackage{booktabs}

\usepackage{url}

\usepackage{caption}
\usepackage{subcaption}

\usepackage{csquotes}

\usepackage[most]{tcolorbox}
\usepackage{xcolor}

\title{SciEx: Benchmarking Large Language Models on Scientific Exams with Human Expert Grading and Automatic Grading}

\author{Tu Anh Dinh, Carlos Mullov, Leonard Bärmann, Zhaolin Li, Danni Liu, Simon Reiß, \\ \textbf{Jueun Lee, Nathan Lerzer, Jianfeng Gao, Fabian Ternava, Tobias Röddiger, } \\\textbf{Alexander Waibel, Tamim Asfour, Michael Beigl, Rainer Stiefelhagen, } \\\textbf{Carsten Dachsbacher, Klemens Böhm, Jan Niehues}\\
  Karlsruhe Institute of Technology, Karlsruhe, Germany\\
  \texttt{\{firstname\}.\{lastname\}@kit.edu}}

\newcommand{\tablefontsize}[0]{\small}

\begin{document}
\maketitle
\begin{abstract}
With the rapid development of Large Language Models (LLMs), it is crucial to have benchmarks which can evaluate the ability of LLMs on different domains. One common use of LLMs is performing tasks on scientific topics, such as writing algorithms, querying databases or giving mathematical proofs. Inspired by the way university students are evaluated on such tasks, in this paper, we propose SciEx - a benchmark consisting of university computer science exam questions, to evaluate LLMs' ability on solving scientific tasks. SciEx is (1) multilingual, containing both English and German exams, and (2) multi-modal, containing questions that involve images, and (3) contains various types of freeform questions with different difficulty levels, due to the nature of university exams. We evaluate the performance of various state-of-the-art LLMs on our new benchmark. Since SciEx questions are freeform, it is not straightforward to evaluate LLM performance. Therefore, we provide human expert grading of the LLM outputs on SciEx. We show that the free-form exams in SciEx remain challenging for the current LLMs, where the best LLM only achieves 59.4\% exam grade on average. We also provide detailed comparisons between LLM performance and student performance on SciEx. To enable future evaluation of new LLMs, we propose using LLM-as-a-judge to grade the LLM answers on SciEx. Our experiments show that, although they do not perform perfectly on solving the exams, LLMs are decent as graders, achieving 0.948 Pearson correlation with expert grading.

\end{abstract}

\section{Introduction}
In recent years, Large Language Models (LLMs) have proven their usefulness across a wide range of tasks, from conversational agents to code generation \cite{rajkumar2022evaluating,abbasian2023conversational, 10.1145/3539618.3594250}.
Given the fast pace of development in the field, with an increasing number of LLMs being trained and released, it is important to have indicators of LLM performance on different domains.
This can be achieved by establishing evaluation benchmarks that assess the capabilities of LLMs across diverse use cases. 

One use case of LLMs is to handle scientific tasks. Some previous works have introduced benchmarks containing questions on science topics  \cite{welbl-etal-2017-SciQ, lu2022ScienceQA, gilson2022medical, schubert2023neurology, zhang2024m3exam}. However, these benchmarks are limited to multiple-choice questions. This restricts the variability of questions, such as instruction-follow ones like "write a mathematical proof for this statement ...". Additionally, it is difficult to ask certain types of questions in a multiple-choice way without including the answer in the question itself. Multiple-choice benchmarks therefore create a gap between testing and actual usage, since they only evaluate whether the LLMs choose the correct answer, whereas in real life, the users are more likely to ask open-ended questions to the LLMs. In contrast, some other works have introduced freeform question benchmarks. These works either convert multiple-choice questions to freeform questions \cite{bhakthavatsalam2021arc-da}, or focus on a specific type of problem such as answering questions related to a paper \cite{dasigi-etal-2021-Qasper}, thus still limiting the variability of the questions.

In this paper, we introduce a new benchmark, termed SciEx (Scientific Exams), designed to evaluate this capability. Inspired by the way students are evaluated in university, we created the benchmark by evaluating the performance of LLMs on university computer science exams. SciEx's questions are in various formats, from multiple choice to open-ended, thus making it suitable to evaluate LLM's capabilities of generating free-text answers that fit the requirements of the questions. It is multilingual, containing exams in both German and English. It is multimodal, as exam questions can also contain figures. The set of questions is a good mix of different difficulty levels since they are designed for university exams. This enables us to evaluate LLMs on different levels, and we found that stronger LLMs tend to perform better on more difficult questions.

Unlike the previous multiple-choice benchmarks, the questions in SciEx are freeform, making it non-trivial how to evaluate the LLM output. Therefore, we make use of expert grading, i.e., having the lecturers grade the LLM output the same way they would grade student answers. We also ask the experts to perform qualitative analysis of the LLM output. With expert grading, we provide a highly reliable way of evaluating LLMs, which is more reliable than previous work that uses crowd-sourced evaluation. Expert grading by lecturers also provides an opportunity to compare LLMs' performance to university student performance in a similar setting. We find that the stronger LLMs, i.e., Claude and GPT-4V, are able to outperform the student average. However, they are still far from perfect, achieving only 59\% across SciEx exams.

Since new LLMs are constantly being released, we cannot fully rely on expert grading for evaluation. Therefore, we provide an automatic grading scheme by using LLM as a judge so that future LLMs can also be evaluated on SciEx. Interestingly, we find that, although LLMs do not perform too well as examinees, they perform well as graders, achieving over 0.948 Pearson correlation to expert grading in the best setting.

In summary, our contributions are as follows:
\begin{itemize}
\setlength{\itemsep}{0pt}
    \item SciEx\footnote{We release SciEx under CC BY-NC-SA 4.0 license. Code: \url{https://github.com/TuAnh23/SciEx}. Data: \url{https://huggingface.co/datasets/tuanh23/SciEx}.} - a freeform, multimodal, multilingual benchmark consisting of university computer science exams, outputs of various LLMs on the exams, and expert grading of the LLM output. 
    \item Detailed quantitative and qualitative analysis comparing LLM to student performance.
    \item Automatic grading with 0.948 Pearson correlation to expert grading
\end{itemize}

\section{Related Work}
\paragraph{General-Purpose LLM Benchmarks} In order to rank different LLMs, there are several commonly used public benchmarks. For example, \citet{zheng2024judging} introduced MT-bench and Chatbot Arena. MT-bench is a multi-turn question set; and Chatbot Arena is a crowdsourced battle platform for LLMs where the users can ask their questions and vote for the better LLM answer. Another benchmark is MMLU \cite{hendrycks2020mmlu}, which is a multitask dataset covering multiple domains such as mathematics, US history and law. 

\paragraph{Scientific LLM Benchmarks} To specifically focus on the scientific domain, previous studies have established benchmarks, such as SciQ \cite{welbl-etal-2017-SciQ} and ScienceQA \cite{lu2022ScienceQA}, which feature questions spanning various scientific subjects. More recent works have focused on benchmarking LLMs on solving exam questions on some narrow science domains such as medical \cite{gilson2022medical} or neurology \cite{schubert2023neurology}. M3Exam \cite{zhang2024m3exam}, in contrast, provides exam questions to benchmark LLMs which span over multiple topics and multiple educational levels (primary, middle, and high school). However, all benchmarks mentioned above are limited to multiple-choice questions. While this simplifies the evaluation process, it does not allow us to assess the LLMs' capability to generate natural text.

Other studies have instead provided scientific benchmarks with open-ended questions. Some examples are Qasper \cite{dasigi-etal-2021-Qasper} and ARC-DA \cite{bhakthavatsalam2021arc-da}. However, Qasper only focuses on questions about NLP papers rather than on general computer science topics. ARC-DA is closer to our work, since it contains open-ended questions taken from science exams and quiz sources. However, these are created by converting questions that were originally multiple-choice, thus not covering certain types of typical freeform questions (e.g. those that require mathematical proofs, or long explanations). 

Different from these works, SciEx is created from university computer science exams, thus naturally providing diversity in the types of questions as well as having freeform format.

\paragraph{Freeform Answer Evaluation} Compared to benchmarks with multiple-choice questions, evaluating LLMs' performance on freeform questions is not straightforward. Similar to evaluation conditions in tasks such as machine translation or summarization, there are multiple correct answers, or multiple ways to express a correct answer for a single input. Therefore, it is insufficient to evaluate a model's output by comparing it to a gold standard answer. Ideally, in these cases, we can evaluate by human judgment. For example, the ARC-DA benchmark \cite{bhakthavatsalam2021arc-da} uses a crowdscoring pipeline for evaluation. Chatbot Arena \cite{zheng2024judging} also uses crowdsourcing, where the users vote between pairs of LLM output. However, human evaluation is inherently non-scalable. Therefore, previous works have used automated metrics. Some traditional metrics such as BLEU \cite{papineni-etal-2002-bleu} or ROUGE \cite{lin-2004-rouge} compare the model's output to some gold-standard answer on the surface level, i.e., word matching. More advanced metrics, such as s BERTScore \cite{bert-score}, BLEURT \cite{sellam-etal-2020-bleurt}, and BARTScore \cite{NEURIPS2021_e4d2b6e6}, are model-based, thus being able to evaluate answers on the semantic level. 

One recent approach is to use LLMs for evaluation, termed \enquote{LLM-as-a-judge}. \citet{liu-etal-2023-g,chiang-lee-2023-large,zheng2024judging} find that, although still prone to biases, LLM-as-a-judge for textual modality has high agreement with human scoring when a strong judge LLM is used. However, when including images, \citet{chen2024mllm} find that the performance of LLM-as-a-judge is no longer as well correlated to human judgment. Nevertheless, LLM-as-a-judge is a promising way to perform scalable evaluation.

In our work, we make use of LLM-as-a-judge for automatic grading of LLM answers on SciEx exams, and find that they have good correlation to human expert grading on both text-only and image-related questions.

\section{The SciEx Benchmark}
The components of SciEx are as follows.

\paragraph{Univeristy Exams} SciEx contains university computer science exams in a unified JSON format. The exams are taken from the following computer science courses at the Karlsruhe Institute of Technology from the 2022/2023/2024 semesters:
\begin{itemize}
\setlength{\itemsep}{0pt}
    \item Natural Language Processing (NLP)
    \item Advanced Artificial Intelligence (AI2)
    \item Deep Learning and Neural Networks (DLNN)
    \item Deep Learning for Computer Vision (DL4CV2)
    \item Human–Computer Interaction (HCI)
    \item Databases (DBS) for the years 2022 and 2023
    \item Computer Graphics (CG)
    \item Theoretical Foundations of Computer Science (TGI)
    \item Algorithms (ALGO)
\end{itemize}
Descriptions of the exams are in Appendix \ref{sec:exam_description}. In total, SciEx contains 10 exams, among which 5 exams are in English and 7 exams are in German (some exams are provided in both languages).

There are in total 154 unique questions. Each question is annotated with (1) the maximum points that can be achieved and (2) a difficulty level among \textit{Easy, Medium, Hard}. Most questions are provided with gold reference answers and average student performance. The detailed per-question statistics are shown in Table \ref{tab:ques_stats}.

\begin{table}[htbp]
\tablefontsize
\centering
\begin{tabular}{lc}
\toprule
                                     & \multicolumn{1}{l}{Question count} \\\hline
Total                                  &    154                           \\
English / German*                      & 95 / 97                              \\
Text-only / Image-related              & 121 / 33                             \\
Easy / Medium / Hard                     & 51 / 71 / 32                           \\
With / Without reference      & 120 / 34                             \\
With / Without student average & 117 / 37                             \\
\bottomrule
\multicolumn{2}{l}{*: Some questions are provided bilingually.}              
\end{tabular}
\caption{Question-level statistics for SciEx.}
\label{tab:ques_stats}
\end{table}

\paragraph{LLM-Generated Answers} SciEx contains answers produced by 7 LLMs on the exam questions. The details of the LLMs are shown in Table \ref{tab:llms}. In Table \ref{tab:llms}, only Llama3 was not used to solve the exam, since it was released at a later point of conducting this paper. In total, we obtained 1120 question-answer pairs.

\begin{table*}[]
\tablefontsize
\centering
\begin{tabular}{lcccc}
\toprule
                     & Full name                                                        & \# Params            & \multicolumn{1}{l}{Quant.} & \multicolumn{1}{l}{Handle Image} \\ \hline
\textbf{Proprietary} & \multicolumn{1}{l}{}                                             & \multicolumn{1}{l}{} & \multicolumn{1}{l}{}             & \multicolumn{1}{l}{}              \\ \cline{1-1}
Claude               & Claude-3-opus-20240229                                           & -                    & -                                & yes                               \\
GPT-4v               & gpt-4-vision-preview                                             & -                    & -                                & yes                               \\
GPT-3.5              & gpt-3.5-turbo-0125                                               & -                    & -                                & no                                \\ \hline
\textbf{Open source} &                                                                  &                      &                                  &                                   \\ \cline{1-1}
Llama3               & \multicolumn{1}{l}{MaziyarPanahi/Meta-Llama-3-70B-Instruct-GGUF} & 70B                  & 4 bit                            & no                                \\
Mixtral              & Mistralai/Mixtral-8x7B-Instruct-v0.1                             & 8x7B                 & 5 bit                            & no                                \\
Qwen                 & Qwen/Qwen-72B                                                    & 72B                  & 2 bit                            & no                                \\
Mistral              & Mistralai/Mistral-7B-Instruct-v0.2                               & 7B                   & -                                & no                                \\
Llava                & Llava-hf/Llava-v1.6-Mistral-7b-hf                                & 7B                   & -                                & yes                              \\
\bottomrule
\end{tabular}
\caption{Details of the LLMs in consideration.}
\label{tab:llms}
\end{table*}

\paragraph{Expert Grading and Automatic Grading} Each question-answer pair is assigned a score by an expert. In order to guide future work to evaluate new LLMs on SciEx without relying on human expert grading, we also provide automatic grading generated by Mixtral, Llama3 and GPT4V.

\subsection{Data Creation}
The data creation process is as follows. 
\paragraph{Exam Collection} We collect university exams from different courses. We additionally ask the lecturers to provide us with the reference answers, the difficulty level of each question, and the average student grades on each question.

\paragraph{Exam Formatting} We convert every exam into a unified JSON format. Each exam includes a list of questions, where each question includes an index, its content, and potentially path to any related images. An example is shown in Appendix \ref{sec:exam_format}.

\paragraph{LLM-Generated Answers} We pass the exams to the LLMs listed in Table \ref{tab:llms} (except Llama3 due to later release), one question at a time. Questions that contain images are handled differently depending on the LLM. For the text-only LLMs, we exclude the images and only pass the question text to the models. For Llava, since it is trained to handle only 1 image at a time, we concatenate the images into one, with blank padding around the images as separators before feeding it to the model. Claude and GPT-4V can take multiple images, however, there is no pre-defined way of referencing the image within the text. In our work, we reference the image by mentioning the image caption within the question text, and add the text caption to the image. 

Since the considered LLMs can only output text, for questions asking to draw on images, we ask the LLMs to describe in text what should be drawn.

The detailed prompts for LLMs to generate the answers are shown in Appendix \ref{sec:do_exam_prompt}.

\paragraph{Expert Grading} We then give the LLM answers back to the lecturers, who proceed with grading the LLM output the same way they would grade student answers. We anonymized the LLMs' names in order to avoid bias during exam grading. We also build a user interface for collecting the grades (see Appendix \ref{sec:streamlit} for more details).

With expert grading, the evaluation of the LLM output is highly reliable. Most importantly, the expert graders are generally the ones who designed the exam questions. We additionally ask the expert graders to provide their comments on the LLM output to further understand LLMs' behaviors when solving the exams.

\subsection{Automatic Grading} In addition to expert grading, we also provide automatic grading using LLM-as-a-judge, so that we can evaluate future LLMs on SciEx. We use the stronger models, i.e., Mixtral, Llama3 and GPT-4V, to conduct the grading. Given a tuple containing question, answer, and maximum score, we ask the LLMs to output a single score between 0 and the maximum. We include reference answers to the grading prompt. We ask the LLMs to provide chain-of-thought reasoning \cite{wei2022cot, chiang-lee-2023-closer} before giving the grade. We also include examples for grading in the prompt, so-called few-shot judge \cite{zheng2024judging}. Each example is a tuple consisting of a question, an answer, and the expert-provided grade. We try out different settings to select the examples, as described below.

Let’s say we want to grade Question M from Exam A, answered by Examinee X. Then the shot examples can be chosen in one of the three ways:

\begin{itemize}
    \item \textbf{Same question}: Select examples from the same Question M from Exam A, but answered by a different Examinee Y. This mimics the real-life scenario where we use the expert resource to grade some answers of the same exam, then use it to guide the LLM graders.
    \item \textbf{Same exam}: select examples from a different Question N from the same Exam A, Answered by a different Examinee Y. Here the examples are in the same domain as the question-answer pair in consideration. This mimics the real-life scenario where, e.g., we have expert grading on exams of the same course from previous years to guide LLM graders.
    \item \textbf{Different exam}: select examples from a different Question N, from a different Exam B, Answered by a different Examinee Y. This mimics the real-life scenario where, e.g., we have expert grading for an exam of another course to guide the LLM graders.
\end{itemize}
Examinee Y and  Question N are chosen randomly. For Exam B, we opt to select the exams that do not heavily require images for simplicity in the prompt.

Intuitively, the example-selection settings above have decreasing levels of relevance to the actual grading query, but increasing easiness to collect. The detailed prompts for LLMs to grade answers are shown in Appendix \ref{sec:grader_prompt}.

\section{Experiments}
In this section, we describe our experiments and results. For prompting the proprietary LLMs, we use their APIs, namely OpenAI\footnote{\url{https://platform.openai.com/}} and Anthropic\footnote{\url{https://console.anthropic.com/}}. For the open-source models, we obtain model checkpoints from the Huggingface\footnote{\url{https://huggingface.co/}} model hub. We perform inference with the LLMs using llama.cpp\footnote{\url{https://github.com/ggerganov/llama.cpp}} with the default sampling strategy. The experiments with open-sourced models are conducted on an NVIDIA RTX A6000 GPU with 48GB VRAM. 

For our analysis, we consider the exam-level and question-level grades. An exam-level grade is the sum of the grades of all questions in the exam.

\subsection{Quantitative Analysis}
We analyze the performance of the LLMs on SciEx with expert grading. For both exam level and question level, we normalize the grade to be between $0$ and $100\%$, since they have different scales. The normalization is done by taking the scores obtained by the examinee divided by the maximum score possible per exam/question, where the maximum scores possible are predefined by the lecturers. 

We also report on the German grade scale. In the German scale, the grades range from $1.0$ to $5.0$, where $1.0$ is the highest grade and $4.0$ is the passing threshold. The detailed mapping from the scores to the German grade scale is defined by the lectures, adjusted based on the overall performance of the students taking the exams. 

We compare the performance of the LLMs to students from different aspects: language, difficulty level, and modality, i.e., questions with or without images.

\subsubsection{General Observations} \label{sec:sciex_challenging}

\paragraph{SciEx is Challenging} The performance of the LLMs on SciEx provided by expert grading is shown in  Table \ref{tab:overall_rank}. The bigger-sized LLMs (Claude, GPT-4V, GPT-3.5, Mixtral and Qwen) can achieve exam passing grades (i.e., grades that are better than $4.0$ in the German scale). However, the best-performing model (Claude) only achieves $59.4\%$ of the maximum points, which is far from perfect.

Compared to the student average, most LLMs have worse performance. Only the strongest proprietary LLMs, i.e., Claude and GPT-4V, can achieve grades that are better than the students'.

\begin{table}[htbp]
\tablefontsize
\centering
\begin{tabular}{lcc}
\toprule
                                                         & Grade (\%) $\uparrow$                                               & German Scale $\downarrow$                                          \\ \hline
\textbf{Proprietary}                                     & \multicolumn{1}{l}{}                                     & \multicolumn{1}{l}{}                                     \\ \cline{1-1}
Claude                                                   & 59.4                                                     & 2.4                                                      \\
GPT-4V                                                    & 58.2                                                     & 2.5                                                      \\
GPT-3.5                                                    & 32.8                                                     & 3.9                                                      \\ \hline
\textbf{Open source}                                     & \multicolumn{1}{l}{}                                     & \multicolumn{1}{l}{}                                     \\ \cline{1-1}
Mixtral                                                  & 41.1                                                     & 3.5                                                      \\
Qwen                                                     & 35.4                                                     & 3.7                                                      \\
Mistral                                                  & 25.9                                                     & 4.2                                                      \\
Llava                                                    & 21.5                                                     & 4.3                                                      \\ \hline
Student avg.                                              & 45.3                                                     & 3.1                                                      \\
\bottomrule
\end{tabular}
\caption{Average performance of LLMs, exam level.}
\label{tab:overall_rank}
\end{table}

\paragraph{SciEx Versus Other Benchmarks} The ranking of the LLMs on SciEx in Table \ref{tab:overall_rank} generally agrees with other public benchmarks.
However, SciEx seems to be more challenging. For example, the best LLM accuracy achieved on MMLU's various tasks is $88.8\%$. The best accuracy achieved on M3Exam multiple choice questions is $72.92\%$. Although these scores are not directly comparable, it indicates that SciEx provides a more challenging test set for future LLMs.

\subsubsection{Influential Factors}

\begin{figure}
     \centering
     \begin{subfigure}[b]{\columnwidth}
         \centering
         \includegraphics[width=\columnwidth]{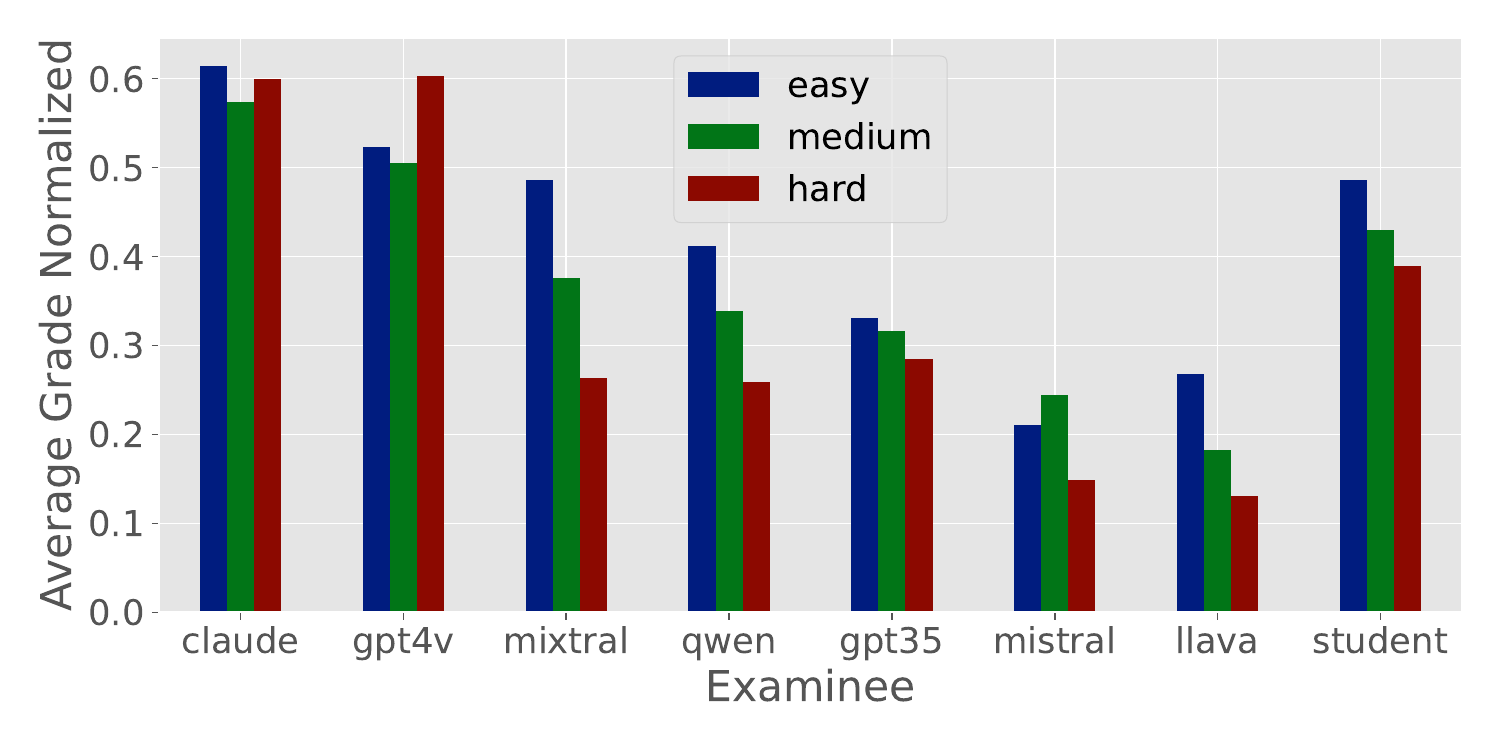}
          \caption{LLMs' and students' scores.}
          \label{fig:difficulty_grade}
     \end{subfigure}
     \hfill
     \begin{subfigure}[b]{\columnwidth}
         \centering
         \includegraphics[width=\columnwidth]{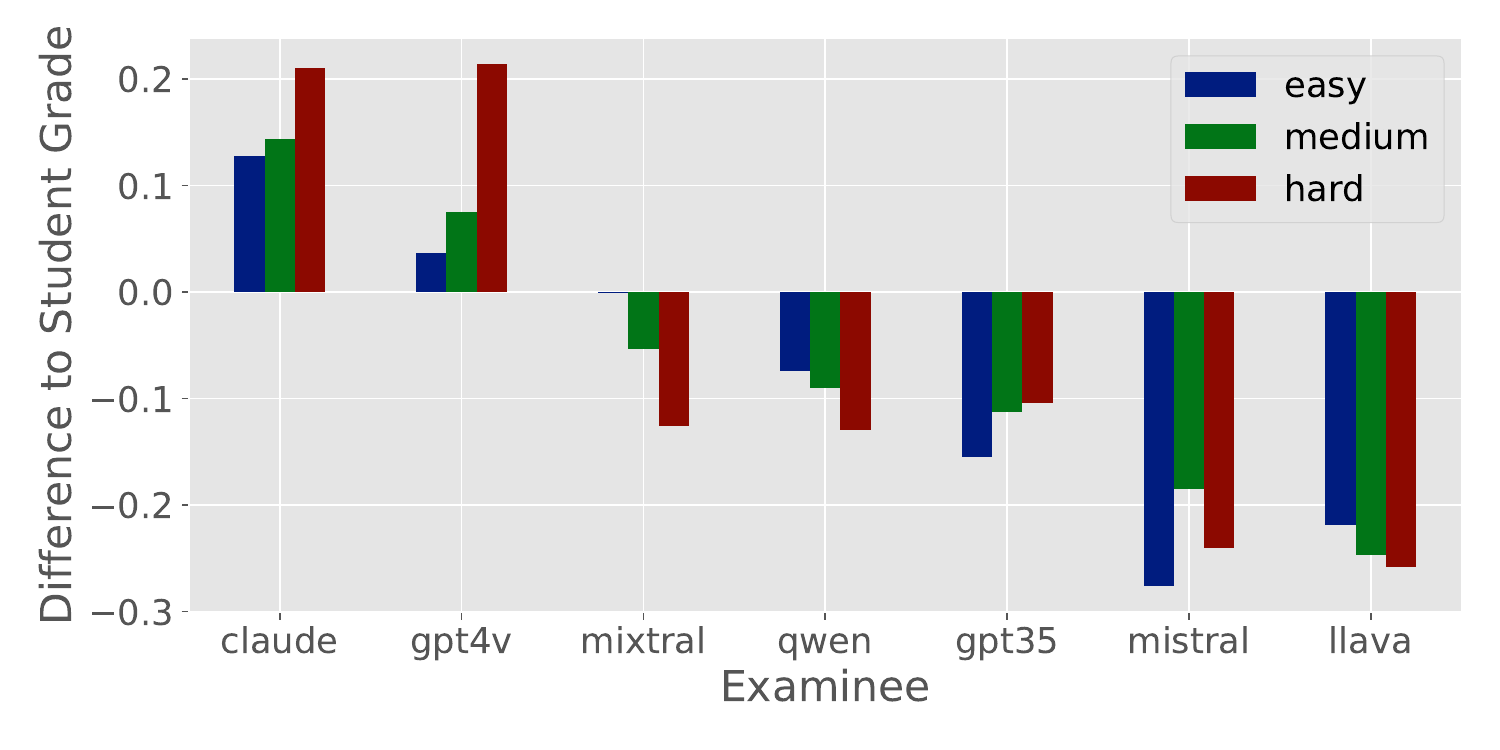}
          \caption{Difference between LLM scores and student scores.}
          \label{fig:difficulty_grade_studdiff}
     \end{subfigure}
    \caption{Question-level scores grouped by difficulty.}
\end{figure}

\paragraph{Difficulty Levels} 
Figure \ref{fig:difficulty_grade} shows the influence of the difficulty level on the examinee grades. As can be seen, the student performance aligns with the difficulty level of the questions: they perform better on easier questions. Some weaker LLMs, e.g., Mixtral, Qwen, GPT-3.5, Llava, align with the students. However, the stronger LLMs, i.e., Claude and GPT-4V, perform better on harder questions. This is an indication that difficulty levels from human perspective do not always align with LLMs' perspective. This is also confirmed by looking at the Pearson correlations between the LLMs' grades and the student average grades on the question level. These correlations are between 0.4 and 0.6, indicating that LLM grades and the student grades are not highly correlated. 

One possible explanation for the mismatch between LLMs performance and question difficulty level could be that, in some exams, there can be some “template questions”, i.e., questions that are repeated over the years, where students can just learn by heart how to systematically solve them. While this would be marked as “easy” by the lecturer, it might not be as easy for the LLMs, since the LLMs are not previously exposed to these “template questions”. Another potential explanation is that math-type \textit{easy} questions are hard for the LLMs, while long-text \textit{hard} questions are easy for them. 

In Figure \ref{fig:difficulty_grade_studdiff}, we plot the difference between LLM scores and student scores. The stronger LLMs, i.e., Claude and GPT-4V, outperform the students the most on hard questions. Weaker LLMs, on the other hand, generally fall behind students the most on hard questions. Looking at each difficulty level independently, we observe that the ranking of the LLMs changes across different levels. This aligns with the findings made by \citet{li2024introducing}, where they show that the LLM rankings change on a subset of evaluation prompts that are artificially labeled as \textit{hard}.

\begin{figure}[t]
  \includegraphics[width=\columnwidth]{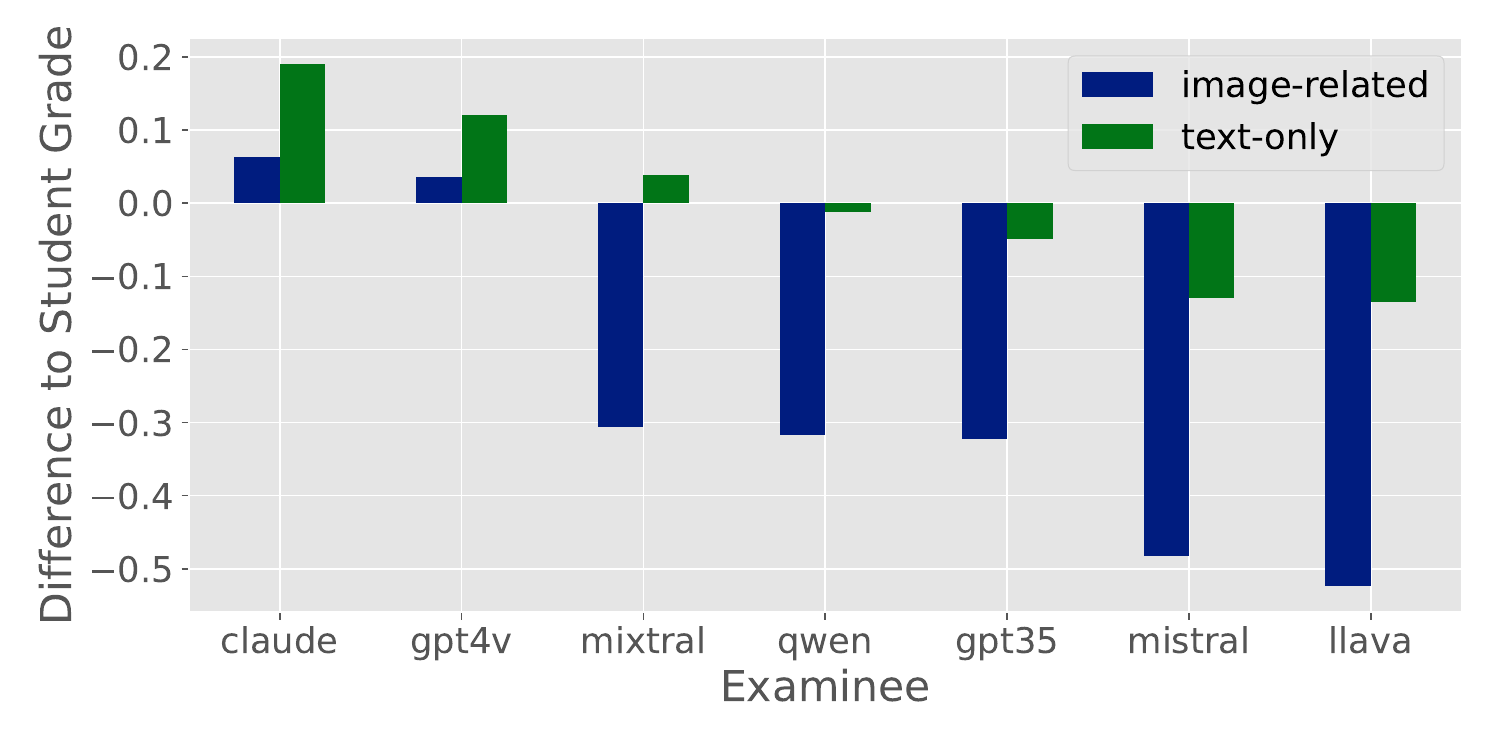}
  \caption{Difference between LLM scores and student scores, question-level, grouped by with/without images. Only Claude, GPT-4V and Llava can handle images.}
  \label{fig:image_grade}
\end{figure}

\paragraph{Text-only versus Image-related Questions} Figure \ref{fig:image_grade} shows the influence of images on the difference between LLM and student scores. Recall that for the text-only LLMs, we exclude the images and only pass the question text to the models. Trivially, the text-only LLMs perform poorly on the image-related questions. The strong, multi-modal LLMs, i.e., Claude and GPT4, outperform the students on both image-related questions and text-only questions, but the performance gap is still larger for text-only questions. Llava, although can handle images, still falls behind student performance by a large margin on image-related questions. This shows that LLMs' image-handling capability is still not as advanced as for text.

\begin{figure}[htbp]
  \includegraphics[width=\columnwidth]{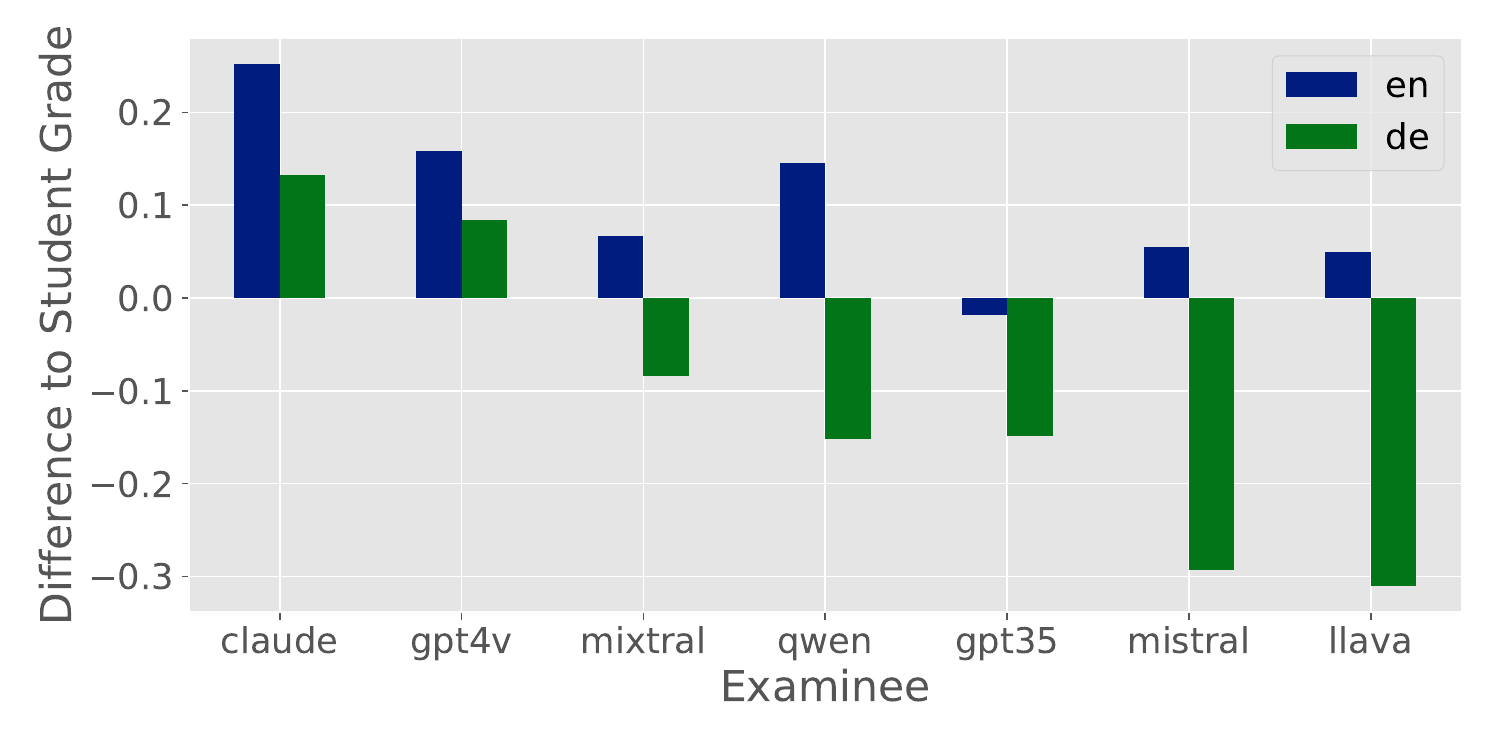}
  \caption{Difference between LLM scores and student scores, question-level, grouped by languages.}
  \label{fig:lang_grade}
\end{figure}

\paragraph{Language}  Figure \ref{fig:lang_grade} shows the influence of languages on the difference between LLM scores and student scores. When the questions are in English, all LLMs, except for GPT-3.5, outperform the student average. However, for German, either the LLMs outperform students by a smaller gap, or fall behind student performance. It can be concluded that LLMs are still superior in English than other languages like German, although German can be considered a high-resource language.

Since some models are not made to deal with images, or with languages other than English, we additionally analyze LLMs' performance on text-only and English-only questions. On this subset of questions, the grades obtained by the models are generally better, and more models would outperform the student average. More details can be found in Appendix \ref{sec:textOnlyEnglishOnly}.

\subsection{Qualitative Analysis}
In this section, we summarize the observations made by the graders while grading the LLMs.

\paragraph{General Behaviours} The graders observed some common behaviors made by the LLMs. Some solutions of the LLMs were good language-wise but low-quality content-wise. For students, good language usually correlates strongly with good content. The LLMs tend to output lengthy answers, since, unlike the students, they do not have a time constraint. Some LLMs even ignore when the question specifies that they should ''answer briefly``. There are also some failure cases, although not frequent: (1) Claude refuses to answer the question with "I apologize, but I do not feel comfortable providing answers related to ..." or (2) some LLMs get stuck in decoding loops. Sometimes, instead of answering the question, LLMs give some text that is (or seems) related to the task; rephrase the task; or describe how a task of this nature may be approached in general.

\paragraph{Knowledge-type Questions} On some exams such as \hyperref[itm:ai2]{AI2}, \hyperref[itm:dl4cv2]{DL4CV2}, \hyperref[itm:dlnn]{DLNN}, \hyperref[itm:cg]{CG}, questions which students can answer by learning the lecture content by heart are quite easy for the LLM. For the \hyperref[itm:dl4cv2]{DL4CV2} exam,  very specific questions about neural network architectures which are covered in our lecture seem to be quite common knowledge in the LLMs, which might be due to those papers being included in the training data. However, for other exams such as \hyperref[itm:hci]{HCI}, the models lacked specific course context, which was important for answering many theoretical and open-ended questions. 

\paragraph{Math-related Ability} The LLMs tend to fail on the math-related questions, even the basic ones. For example, they miscount the number of words in a piece of text, or have trouble comparing numbers. For questions that require writing mathematical proof in the \hyperref[itm:tgi]{TGI} exam, all LLMs except for GPT-4V and Claude failed. For GPT-4V and Claude, they are able to pass the \hyperref[itm:tgi]{TGI} exam. Their mistakes are more in line with those that students would make. That is, they are often not successful when making actual proof, and the points where the proof breaks sometimes are the same as the students. Even the better models handle simple geometry questions poorly and/or struggle to follow the instructions of a simple algorithm.

\paragraph{Reasoning Ability} The LLMs do not perform well on questions that require deep thinking and reasoning. For questions of the type “is this statement true or false; reason for your solution”, the LLMs often said “true” and then just repeated the statement or reasoned for the opposite of their claim. This is a similar behavior often seen in students. Sometimes they make self-contradicting arguments: making a statement and then providing arguments for the other side. 

\paragraph{Image Handling} GPT-4V, Claude and Llava can handle images. However, only GPT-4V and Claude have reasonable performance. When the question is about drawing on top of the figures, sometimes the LLMs successfully describe in words what needs to be drawn, but occasionally they just hallucinate a non-existing figure file path.

\subsection{Automatic Grading}

In this section, we evaluate the performance of LLM-as-a-judge approach to automatic grading. We use the expert grades as the gold standard to evaluate automatic graders. We use Pearson correlation on the normalized scores as our metric. Since the LLMs are asked to provide the scores on the same scale as the expert scores, we also provide the Root Mean Squared Error (RMSE) on the originally-scaled scores as a secondary metric. Note that RMSE would correctly put more weight on the questions that have more points, however, it is not as easily interpretable as the Pearson correlation. Therefore, we only report RMSE in Appendix \ref{sec:rmse}. The main results are discussed as follows.

\subsubsection{General}

\begin{table*}[t]
\tablefontsize
\centering
\renewcommand{\arraystretch}{1.2}
\begin{tabular}{ll|ccc|ccc}
\toprule
        &        & \multicolumn{3}{c|}{without ref}                                                                   & \multicolumn{3}{c}{with ref}                                                                      \\
        &        & \multicolumn{1}{l}{same question} & \multicolumn{1}{l}{same exam} & \multicolumn{1}{l|}{diff exam} & \multicolumn{1}{l}{same question} & \multicolumn{1}{l}{same exam} & \multicolumn{1}{l}{diff exam} \\ \hline
\multicolumn{8}{l}{\textbf{Text-only questions}}                                                                                                                                                                         \\
Mixtral & 0 shot & 0.232                             &                               &                               & 0.311                             &                               &                               \\
        & 1 shot & 0.352                             & 0.377                         & 0.364                         & 0.395                             & 0.333                         & 0.275                         \\
        & 2 shot & 0.395                             & 0.299                         & 0.316                         & 0.398                             & 0.271                         & 0.255                         \\
Llama3  & 0 shot & 0.452                             &                               &                               & 0.603                             &                               &                               \\
        & 1 shot & 0.573                             & 0.547                         & 0.500                         & 0.672                             & 0.581                         & 0.645                         \\
        & 2 shot & 0.598                             & 0.522                         & 0.546                         & 0.644                             & 0.596                         & 0.575                         \\
GPT-4V   & 0 shot & 0.607                             &                               &                               & 0.696                             &                               &                               \\
        & 1 shot & 0.605                             & 0.679                         & 0.616                               & 0.653                             & 0.693                         & 0.701                              \\
        & 2 shot & 0.672                             & 0.648                         & 0.674                               & 0.717                             & 0.727                         & 0.678                               \\ \hline
\multicolumn{8}{l}{\textbf{Image-related questions}}                                                                                                                                                                     \\
GPT-4V   & 0 shot & 0.677                             &                               &                               & 0.539                             &                               &                               \\
        & 1 shot & 0.640                             & 0.661                         & 0.611                         & 0.642                             & 0.539                         & 0.749                         \\
        & 2 shot & 0.613                             & 0.632                         & 0.673                         & 0.712                             & 0.465                         & 0.696                        
\\
\bottomrule
\end{tabular}

\caption{LLM grading's Pearson correlation to expert grading on the question level. Note that there are only single scores for zero-shot, since they do not have different shot settings.}
\label{tab:pearson_question_level}
\end{table*}

\paragraph{LLMs Perform Well as Graders} On the exam level, LLM-as-a-judge performs well for automatic grading. The best Pearson correlation to expert grading on the exam level, at 0.948, is achieved by GPT-4V. The open-source Llama3 achieves 0.883 Pearson correlation to expert grading. 

The LLM ranking based on average exam-level grades provided by the GPT-4V grader in comparison to expert grading is shown in Table \ref{tab:ranking}. As can be seen, the ranking is quite identical, except for Mixtral and Qwen's positions being swapped.

\begin{table}[htbp]
\begin{tabular}{lc|lc}
\multicolumn{2}{l|}{\underline{\textbf{Expert grader}}}  & \multicolumn{2}{l}{\underline{\textbf{GPT-4V grader}}}  \\
Examinee & Avg. grade                        & Examinee & Avg. grade                       \\
         & \multicolumn{1}{l|}{(\%, sorted)} &          & \multicolumn{1}{l}{(\%, sorted)} \\
Claude   & 59.4                              & Claude   & 57.7                             \\
GPT-4V   & 58.2                              & GPT-4V   & 56.2                             \\
Mixtral  & 41.1                              & Qwen     & 42.0                             \\
Qwen     & 35.4                              & Mixtral  & 38.2                             \\
GPT-3.5  & 32.8                              & GPT-3.5  & 38.0                             \\
Mistral  & 25.9                              & Mistral  & 24.6                             \\
Llava    & 21.5                              & Llava    & 24.2                            
\end{tabular}

\caption{LLM examinees ranking with expert grader and GPT-4V grader.}
\label{tab:ranking}

\end{table}

The high correlations between expert grading and LLM-as-a-judge grading indicate that, although being far from perfect in solving SciEx exams (discussed in Section \ref{sec:sciex_challenging}), the stronger LLMs are quite reliable for grading the exams. This is useful since we would have to rely less on expert grading to evaluate newly developed LLMs' performance on SciEx. The details of graders' performance under different settings on the exam level are in Appendix \ref{sec:exam_level}.

On the question level, the performance of LLM-as-a-judge is shown in Table \ref{tab:pearson_question_level}. The highest Pearson correlation to expert grading achieved by the LLMs is now around 0.7, which is lower than on the exam level, but still quite high. Surprisingly, the performance of GPT-4V on grading image-related questions is quite comparable to grading text-only questions. This contradicts the finding made by \citet{chen2024mllm}. This could potentially be due to the small number of image-related questions in SciEx, thus the results might not be generalizable.

\paragraph{Few-shot and References Help} The performance of the graders on the question level is shown in Table \ref{tab:pearson_question_level}.
We observe that adding examples (shots) and adding reference answers in the prompt generally increases the performance of the LLM graders. GPT-4V is the strongest grader, followed by Llama3 and Mixtral. This shows that proprietary LLMs are still stronger as judges, aligning with previous studies \cite{zheng2024judging}.

\subsubsection{Grader-specific Behaviours}

\paragraph{Mixtral Grader Tends to Give Full Points} As can be seen from Table \ref{tab:pearson_question_level}, Mixtral has the worst performance on grading the exams. We observe that Mixtral tends to give full points to the answers. Without reference and without examples (0-shot), the portion of answers where Mixtral outputs full points is 67.6\%, significantly higher than Llama3 and GPT-4V, at 19.1\% and 15.1\%, respectively. As a result, Mixtral's precision on giving full points, at 0.181, is much lower than Llama3 and GPT-4V, at 0.380 and 0.527 respectively. As we add more examples and/or add the reference answer to the prompt, the problem is lessened. More details can be found in Appendix \ref{sec:grader_full}.

\paragraph{Mixtral and GPT-4V Copy Grade of Example} For Mixtral and GPT-4V graders, when having one example (shot) from the same question in the prompt without reference, the performance is worse than having the example from the same exam or from a different exam. We hypothesize that this is due to these graders tend to copy the grades of the examples when having a chunk of duplicated text (i.e., the question description) in the example. This is verified when looking at the statistics: Mixtral and GPT-4V copy the grade of the example ~25\% of the time, whereas Llama3 does it 13\% of the time. As a result, Llama3 can best make use of examples from the same question. The problem is reduced when having more than 1 shot or when the reference answer is included.

\subsubsection{Influential Factors}
\paragraph{Different Examinees} As can be seen in Table \ref{tab:grader_on_examinee}, GPT-4V grader has better performance than others, but is more inconsistent: it does worse on grading some LLMs, especially Claude. This is potentially due to Claude being a better examinee than GPT-4V itself, as shown in Section \ref{sec:sciex_challenging}. When using the scores from GPT-4V grader to rank the LLMs, we find that, without reference answer, GPT-4V always ranks itself higher than Claude. This emphasizes the importance of reference answers for grading, especially when the grader is weaker than the examinee.

\begin{table}[htbp]
\tablefontsize
\centering
\begin{tabular}{lccc}
\toprule
        & \multicolumn{3}{c}{Graders} \\ \cline{2-4} 
        & Mixtral  & Llama3  & GPT-4V \\
Claude  & 0.304    & 0.460   & 0.482  \\
GPT-4V   & 0.353    & 0.528   & 0.612  \\
Mixtral & 0.251    & 0.472   & 0.564  \\
Qwen    & 0.351    & 0.556   & 0.736  \\
GPT-3.5   & 0.333    & 0.522   & 0.697  \\
Mistral & 0.291    & 0.467   & 0.601  \\
Llava   & 0.387    & 0.716   & 0.812  \\
\bottomrule
\end{tabular}
\caption{LLM graders performance (i.e., Pearson correlation to expert grading) on different examinees. 
}
\label{tab:grader_on_examinee}
\end{table}

\paragraph{Difficulty Levels} Looking at Table \ref{tab:grader_on_difflevel}, the weaker graders, i.e., Mixtral and Llama3, perform better on grading easier questions. In contrast, GPT-4V performs better in grading harder questions. 

\begin{table}[htbp]
\tablefontsize
\centering
\begin{tabular}{lccc}
\toprule
       & \multicolumn{3}{c}{Graders} \\ \cline{2-4} 
       & Mixtral  & Llama3  & GPT-4V \\
Easy   & 0.374    & 0.602   & 0.628  \\
Medium & 0.293    & 0.524   & 0.690  \\
Hard   & 0.224    & 0.496   & 0.732 \\
\bottomrule
\end{tabular}
\caption{LLM grader's performance (i.e., Pearson correlation to expert grading) on different difficulty levels. 
}
\label{tab:grader_on_difflevel}
\end{table}

\section{Conclusion}
In this paper, we proposed SciEx - a benchmark consisting of scientific university exams, along with expert grading and automatic grading, to evaluate the abilities of LLMs on science topics. SciEx is multilingual, multi-modal, and contains a variety of free-form questions. Our experiments show that SciEx is still quite challenging for current LLMs, where the best LLM can only achieve $59.4\%$ of the exam score on average. Despite that, the LLMs perform well as graders, achieving 0.948 Pearson correlation to the expert grades. This is a promising observation, since we can use strong LLMs for automatic grading of new LLM examinees on SciEx, rather than relying on expert grading. We encourage the research community as well as LLM developers and users to make use of SciEx for evaluating LLMs' scientific capabilities.

\section*{Limitations}
There are certain biases that can occur for SciEx. Firstly, the LLMs do not have time pressure. Therefore, they can output longer answers, which helps them get better grades, as there is a higher likelihood that something will be correct. Secondly, the grading process can not be fully anonymized. It is not easy to mix the LLM answers with student answers for the lecturer to grade, since student answers are usually handwritten. Additionally, the LLMs’ answers content itself might also be easily distinguishable from the students’, since the LLMs tend to, e.g., give longer answers or repeat the questions. Therefore, the lecturers know when they are grading an LLM, thus can bias the score they give. Thirdly, the comparison between the LLMs and the students might be unfair, since the students studied the centralized course material specifically for the exams, while this is not the case for the LLMs. Lastly, due to the reliance on expert resources, the size of SciEx is quite small compared to other scientific benchmarks.

\section*{Ethics}
Our work makes use of student statistics to compare against LLMs' performance. However, we only use the average of the student grades, without disclosing any individual student's information. The student answers are never directly used, as we only ask for the average graders from the lecturers.

Automatic grading, regardless of the high correlation to expert grading, can still be imperfect. We are not suggesting to use LLMs to evaluate students, but to evaluate new models coming out when it is not possible to do human evaluation.

Regarding data consent, we had group meetings and email exchanges to come to an agreement from all lecturers that the data would be made public under the CC BY-NC-SA 4.0 license.

\section*{Acknowledgments}
This work was supported by the Helmholtz Programme-oriented Funding, with project number 46.24.01, project name AI for Language Technologies. It was also supported by funding from the pilot program Core-Informatics of the Helmholtz Association (HGF).

We thank the lecturers for their contribution during the creation of the dataset: Kunyu Peng, Alexander Jaus, David Schneider, Ruiping Liu, Zdravko Marinov, Yufan Chen, Miklós Borsi, Florian Kalinke, Federico Matteucci, Fabian Richter, Bela Böhnke, Jose Cribeiro-Ramallo, Daniel Ebi, Florian Kalinke, Adrian Feilhauer, Wendy Yi, Laura Merker, Miriam Goetze, Jean-Pierre von der Heydt, Max Göttlicher, Thomas Bläsius, Marcus Wilhelm, Michael Zündorf

\bibliography{custom}

\appendix
\clearpage

\section{Exam Description}
\label{sec:exam_description}
The overall description of each exam in SciEx is as follows:
\begin{enumerate}
    \item \label{itm:nlp}    Natural Language Processing (NLP): exam contains questions about word and sequence representation, language modeling, and pretrained models.
    \item \label{itm:ai2}    Advanced Artificial Intelligence (AI2): exam contains questions about natural language processing, signal processing, automatic speech recognition and cognitive robotics.
    \item \label{itm:dlnn}   Deep Learning and Neural Networks (DLNN): exam contains questions about neural network fundamentals, in-depth questions about multihead self attention and calculation questions on backpropagation. 
    \item \label{itm:dl4cv2} Deep Learning for Computer Vision (DL4CV2): exam contains questions about semi-supervised learning, weakly supervised learning, multi-modal text-image models, continual learning, representation learning, interactive segmentation, transfer learning and generative models from recent literature.
    \item \label{itm:hci} Human–Computer Interaction (HCI): exam encompasses fundamental HCI subjects like observational studies, human perception and information processing, user studies, and system design and design analysis. It requires students to utilize theoretical knowledge and perform brief analyses based on the given context.
    \addtocounter{enumi}{1}%
    \edef\labelenumi{\theenumi/\noexpand\theenumi.}%
    \addtocounter{enumi}{-1}%
    \addtocounter{enumi}{1}%
    \item \label{itm:dbs}    Databases (DBS): 2 exams from 2022 and 2023, containing questions about ER (Entity Relationship) modeling, SQL writing and comprehension, relational algebra, and transaction management.
    \def\labelenumi{\theenumi.}
    \item \label{itm:cg}     Computer Graphics (CG): exam contains questions about color and perception, raytracing, shading, data structures, transformations, textures, OpenGl, blending, shaders, procedural modeling, and bezier curves.
    \item \label{itm:tgi}    Theoretical Foundations of Computer Science (TGI): exam contains questions about finite automata, regular languages, push down automata, grammars (Chomsky hierarchy), Turing machine, formal languages, NP-completeness, approximation algorithms, decidability. Most questions require writing mathematical proofs.
    \item \label{itm:algo}   Algorithms (ALGO): exam contains questions on writing proofs (correctness of algorithms, asymptotic run time analysis), knowing basic algorithms and data structures, designing simple new algorithms, selecting the right data structure or algorithm for a given task at hand.
\end{enumerate}

\section{Exam Formatting}
\begin{figure*}[t]
\centering
    \subfloat[Exam in original PDF format.]{
        \includegraphics[width=\columnwidth]{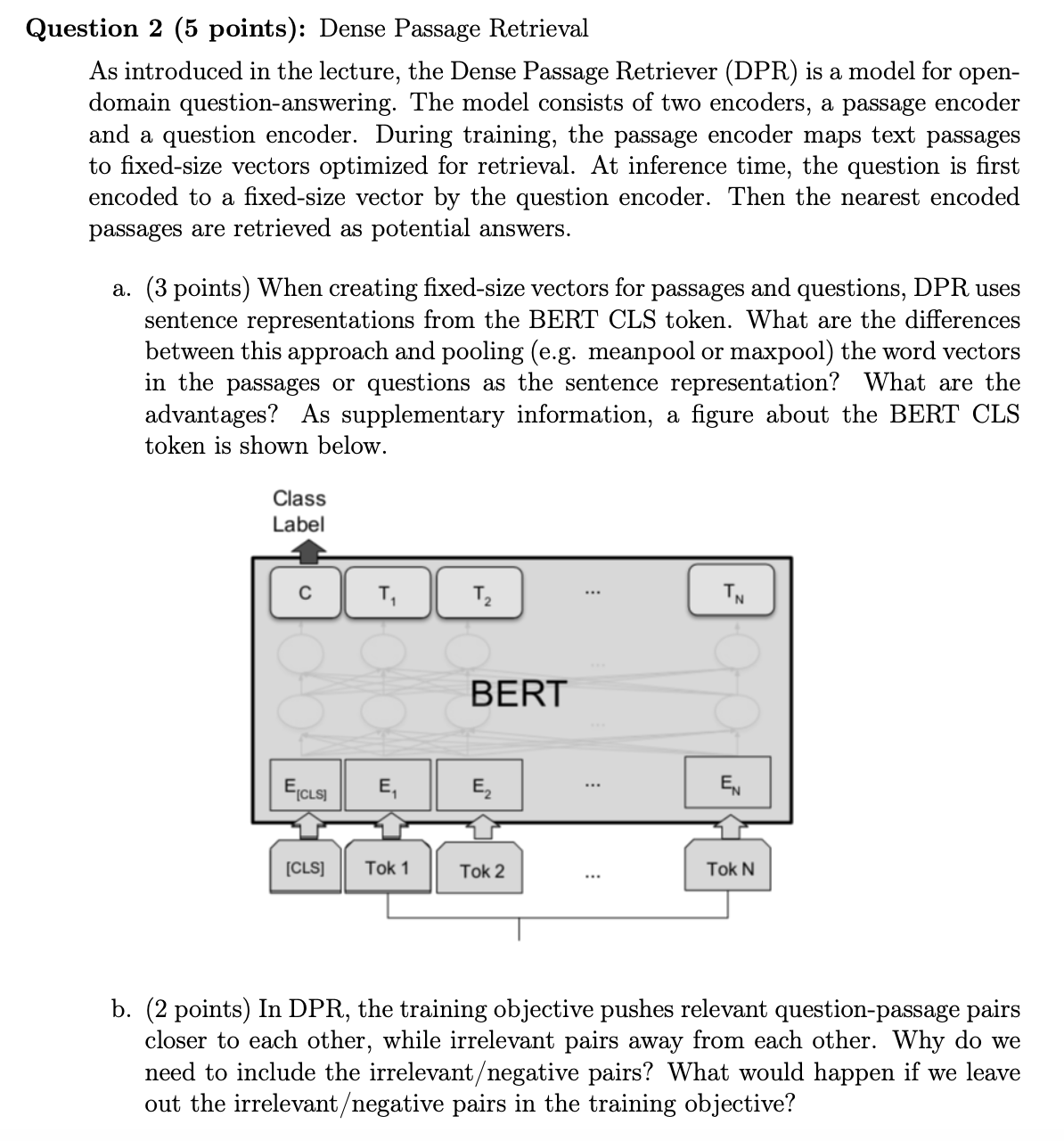}
        \label{fig:pdfques}
    } 
    \hfill
    \subfloat[Exam converted to JSON format.]{
        \includegraphics[width=\columnwidth]{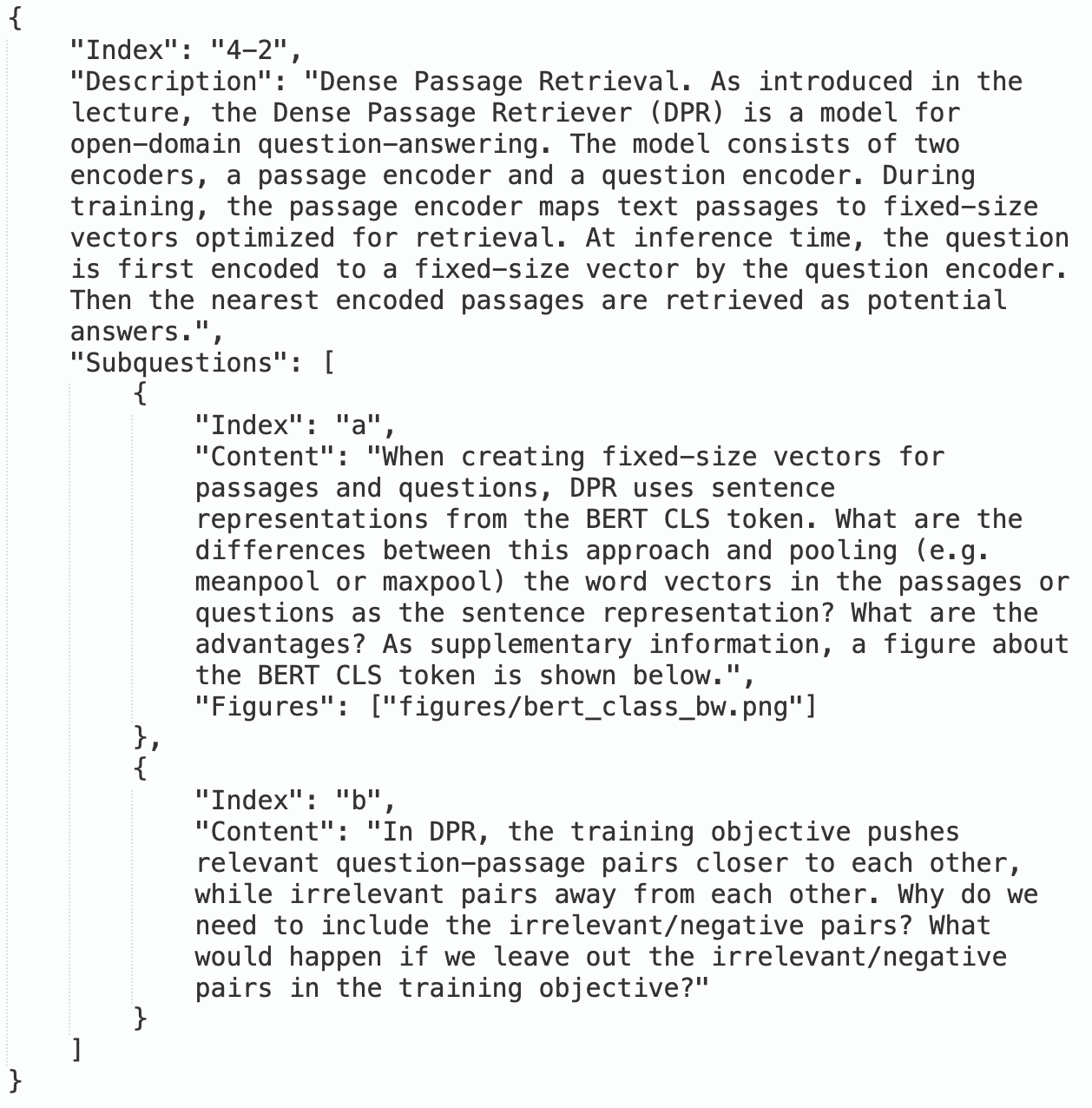}
        \label{fig:jsonques}
    }
    \caption{Exam question before and after being converted to JSON format.}
    \label{fig:exam_format}
\end{figure*}

Originally, exams were in different formats, depending on their creator. We convert the exams into JSON format, with file paths to images if any. An example is shown in Figure \ref{fig:exam_format}.

\label{sec:exam_format}

\section{LLM Answer Generation Prompts} \label{sec:do_exam_prompt}
We provide the prompt in the same language as the exam question to the LLMs to generate answers. The English prompt is shown in Figure \ref{fig:answer_prompt_en}, and the German prompt is shown in Figure \ref{fig:answer_prompt_de}.

\begin{figure*}
\begin{tcolorbox}[colback=blue!5!white, colframe=blue!75!black, sharp corners=southwest]
You are a university student. Please answer the following JSON-formatted exam question.

The subquestions (if any) are indexed.
The provided figures (if any) each contains its path at the bottom, which matches the path provided in the JSON.\\ 

Please give the answers to the question and subquestions that were asked, and index them accordingly in your output.

You do not have to provide your output in the JSON format.
If you are asked to draw on the figure, then describe with words how you would draw it.

Please provide all answers in English.\\

Here is the question: 

\textcolor{blue}{<input text>}
\end{tcolorbox}
\caption{Answer generation prompt in English.}
\label{fig:answer_prompt_en}
\end{figure*}

\begin{figure*}
\begin{tcolorbox}[colback=blue!5!white, colframe=blue!75!black, sharp corners=southwest]
Sie sind Student. Bitte beantworten Sie die folgende JSON-formatierte Prüfungsfrage. 

Die Unterfragen (falls vorhanden) sind indiziert.
Die bereitgestellten Abbildungen (falls vorhanden) enthalten jeweils unten ihren Pfad, der mit dem im JSON bereitgestellten Pfad übereinstimmt. \\

Bitte geben Sie die Antworten auf die gestellten Fragen und Unterfragen an und indizieren Sie diese in Ihrer Ausgabe entsprechend.

Sie müssen Ihre Ausgabe nicht im JSON-Format bereitstellen.
Wenn Sie aufgefordert werden, auf der Figur zu zeichnen, beschreiben Sie mit Worten, wie Sie sie zeichnen würden.

Bitte geben Sie alle Antworten auf Deutsch an. \\

Hier ist die Frage: 

\textcolor{blue}{<input text>}
\end{tcolorbox}
\caption{Answer generation prompt in German.}
\label{fig:answer_prompt_de}
\end{figure*}

\section{LLM Grader Prompts} \label{sec:grader_prompt}

We provide the prompt in the same language as the exam question to the LLMs to perform automatic grading. The English prompt is shown in Figure \ref{fig:grade_prompt_en}, and the German prompt is shown in Figure \ref{fig:grade_prompt_de}.

\begin{figure*}
\begin{tcolorbox}[colback=blue!5!white, colframe=blue!75!black, sharp corners=southwest]
You are a university professor. Please grade the following exam question.\\
The exam question, examinee's answer, correct answer, and the maximum possible score are provided in the format:

\begin{itemize}
    \item \texttt{[question]} <exam\_question> \texttt{[/question]} 
    \item \texttt{[answer]} <answer> \texttt{[/answer]} 
    \item \texttt{[correct\_answer]} <correct\_answer> \texttt{[/correct\_answer]} 
    \item \texttt{[max\_score]} <max\_score> \texttt{[/max\_score]} 
\end{itemize}

The question is provided in JSON format, but the answer can be freeform text. The provided figures in the question (if any) each contain its path at the bottom, which matches the path provided in the JSON. The answer is text-only. If the question asks to draw on the figure, then the answer should contain a text description of how the drawing should be.\\

Please provide the grade between \texttt{[0, <max\_score>]}. Please provide the reasoning for your grade. Please provide your output in the format: 
\begin{itemize}
    \item \texttt{[reason]} <reasoning> \texttt{[/reason]} 
    \item \texttt{[grade]} <grade> \texttt{[/grade]} 
\end{itemize}

Below you are provided with examples on how to perform the grading: 

\textcolor{blue}{<example text>}\\

Here is your input: 

\textcolor{blue}{<input text>}
\end{tcolorbox}
\caption{Grading prompt in English.}
\label{fig:grade_prompt_en}
\end{figure*}

\begin{figure*}
\begin{tcolorbox}[colback=blue!5!white, colframe=blue!75!black, sharp corners=southwest]
Sie sind Universitätsprofessor. Bitte bewerten Sie die folgende Prüfungsfrage.\\
Die Prüfungsfrage, die Antwort des Prüflings, die richtige Antwort und die maximal mögliche Punktzahl werden im Format bereitgestellt::

\begin{itemize}
    \item \texttt{[question]} <Prüfungsfrage> \texttt{[/question]} 
    \item \texttt{[answer]} <Antwort> \texttt{[/answer]} 
    \item \texttt{[correct\_answer]} <korrekteAntwort> \texttt{[/correct\_answer]} 
    \item \texttt{[max\_score]} <maxPunkt> \texttt{[/max\_score]} 
\end{itemize}

Die Frage wird im JSON-Format bereitgestellt, die Antwort kann jedoch Freiformtext sein. Die bereitgestellten Abbildungen in der Frage (falls vorhanden) enthalten jeweils unten ihren Pfad, der mit dem im JSON bereitgestellten Pfad übereinstimmt. Die Antwort ist nur Text. Wenn es sich bei der Frage darum handelt, auf der Abbildung zu zeichnen, sollte die Antwort eine Textbeschreibung darüber enthalten, wie die Zeichnung aussehen soll.\

Bitte geben Sie die Note zwischen [0, <maxPunkt>] an. Bitte begründen Sie Ihre Note. Bitte geben Sie Ihre Ausgabe im Format an: 
\begin{itemize}
    \item \texttt{[reason]} <Grundsatz> \texttt{[/reason]} 
    \item \texttt{[grade]} <Note> \texttt{[/grade]} 
\end{itemize}

Nachfolgend finden Sie ein Beispiel für die Durchführung der Benotung: 

\textcolor{blue}{<example text>}\\

Hier ist Ihre Eingabe: 

\textcolor{blue}{<input text>}
\end{tcolorbox}
\caption{Grading prompt in German.}
\label{fig:grade_prompt_de}
\end{figure*}

\section{User Interface for Expert Grading}
\label{sec:streamlit}
\begin{figure*}[t]
  \includegraphics[width=\textwidth]{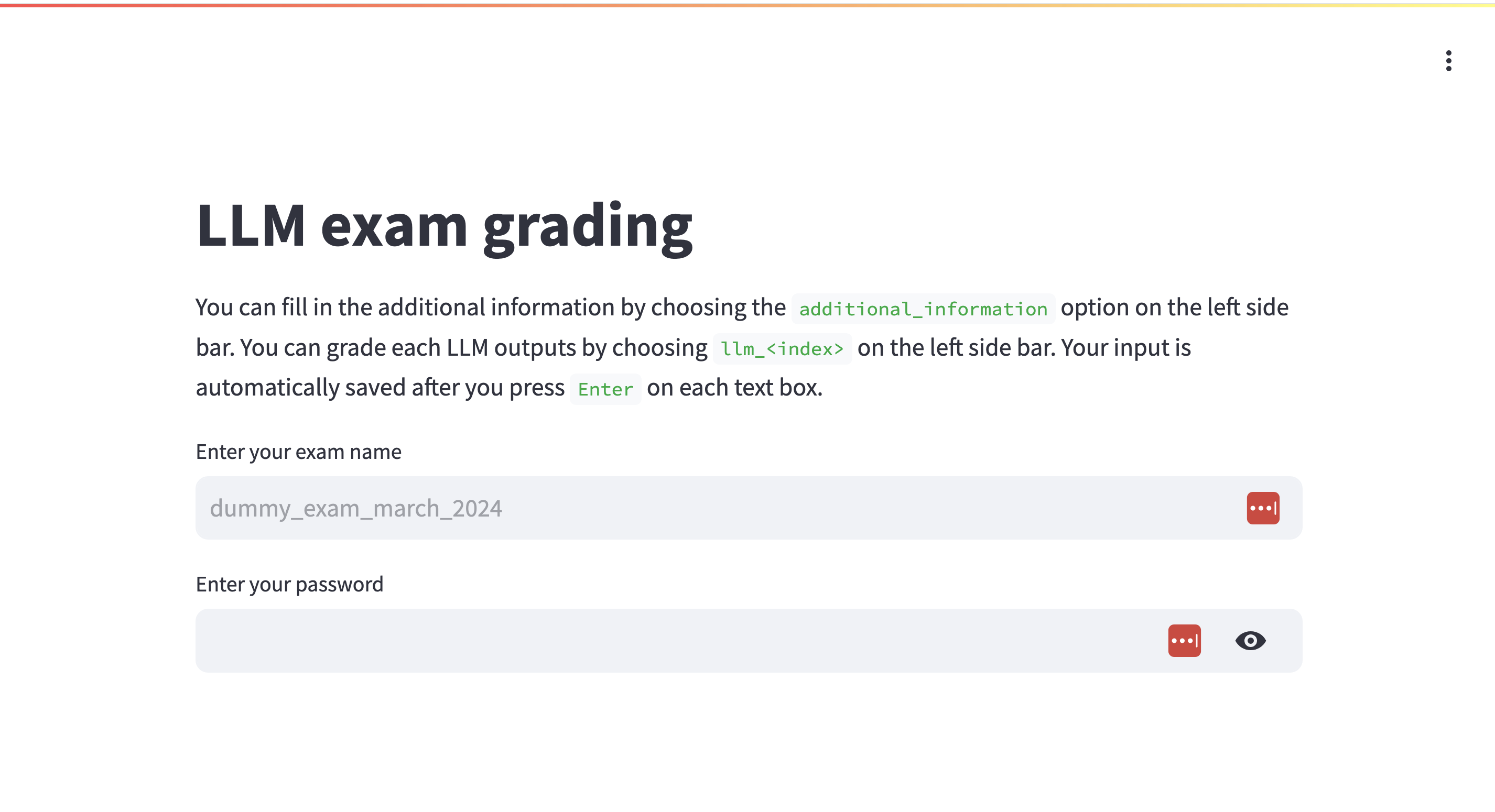}
  \caption{Open page of the grading UI.}
  \label{fig:streamlit_open}
\end{figure*}

\begin{figure*}[t]
  \includegraphics[width=\textwidth]{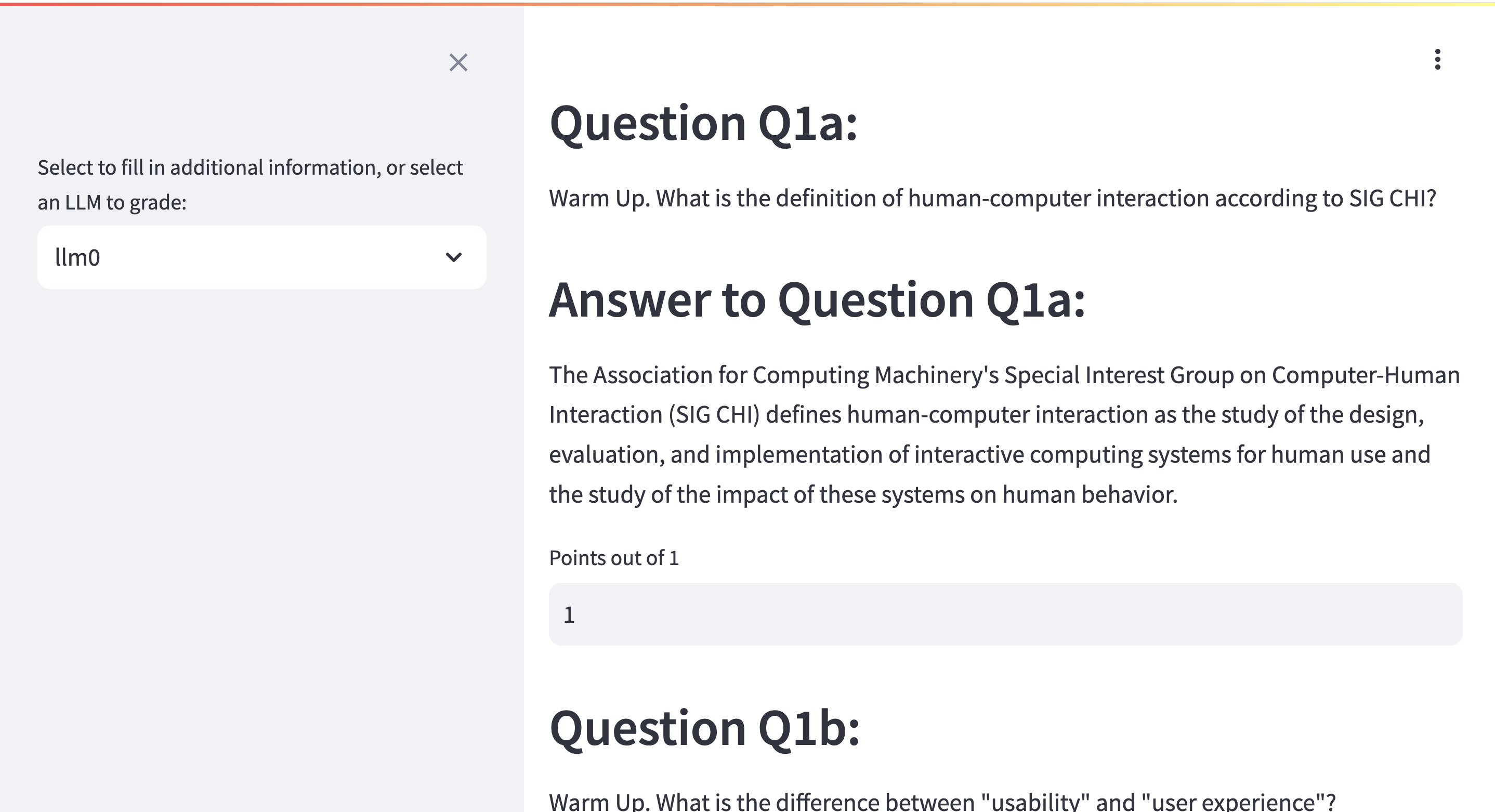}
  \caption{Grading UI showing the question, the answer and textbox to input the grade.}
  \label{fig:streamlit_grade}
\end{figure*}

\begin{figure*}[t]
  \includegraphics[width=\textwidth]{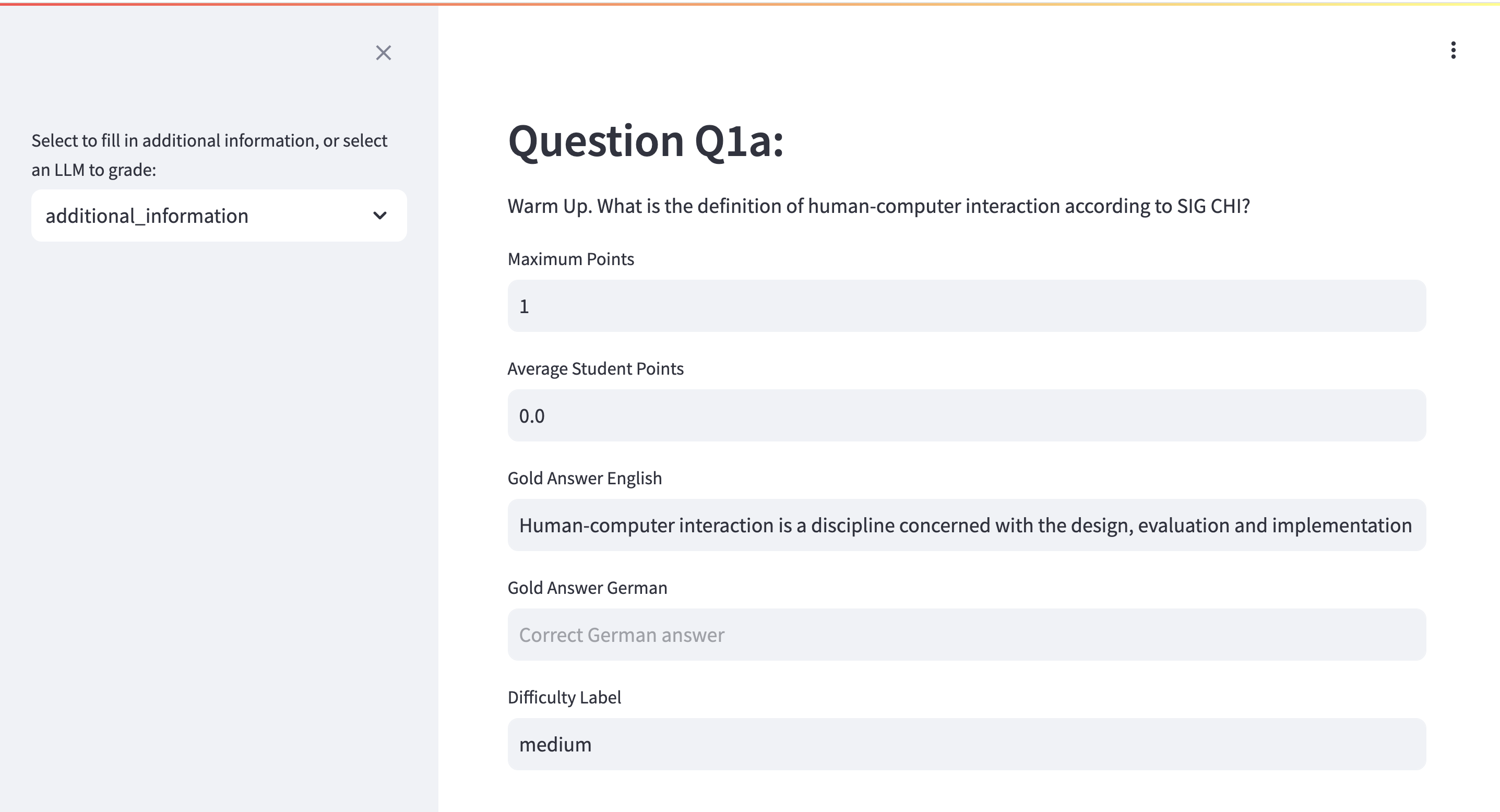}
  \caption{Grading UI where grader can fill in additional information about the question.}
  \label{fig:streamlit_additionalinfo}
\end{figure*}

We instructed the expert grader to use our user interface (UI) for grading. Figure \ref{fig:streamlit_open} shows the open page of the UI, where the grader can choose their exam and enter their password. Figure \ref{fig:streamlit_grade} shows the page for the grading, where the expert is shown with the question, the LLM answer to the question, and a text box to enter the grade. The expert can choose the examinee to grade from the dropdown on the left-hand side. Figure \ref{fig:streamlit_additionalinfo} shows the page to enter additional information about the exam questions, including the maximal achievable score, average student performance, gold answer, and difficulty level.

Once the data is collected, we also ask the experts and have their consent to make the data public.

\section{Performance on Text-only, English-only Questions} \label{sec:textOnlyEnglishOnly}
The performance of the LLMs on text-only, English-only questions is shown in Table \ref{tab:overall_rank_toen}. Qn this subset of questions, besides GPT-4V and Claude, we can see that Mixtral and Qwen also have better performance than the student average.

\begin{table}[htbp]
\tablefontsize
\centering
\begin{tabular}{lcc}
\toprule
                     & Grade (\%)           & German Scale         \\ \hline
\textbf{Proprietary} & \multicolumn{1}{l}{} & \multicolumn{1}{l}{} \\ \cline{1-1}
GPT-4V               & 70.8                 & 1.4                  \\
Claude               & 69.2                 & 1.6                  \\
GPT-3.5              & 47.8                 & 2.9                  \\ \hline
\textbf{Open source} &                      &                      \\ \cline{1-1}
Mixtral              & 61.2                 & 2.0                  \\
Qwen                 & 56.8                 & 2.4                  \\
Mistral              & 48.0                 & 3.2                  \\
Llava                & 42.4                 & 3.5                  \\ \hline
Student avg.         & 56.5                 & 2.4                 \\
\bottomrule
\end{tabular}
\caption{Average performance of the LLMs on the exam level, provided by expert grading, text-only and English-only questions.}
\label{tab:overall_rank_toen}
\end{table}

\section{Grader Performance}
\subsection{Pearson Correlation on Exam Level} \label{sec:exam_level}
The performance of LLM-as-a-judge for automatic grading on the exam level is shown in Table \ref{tab:pearson_exam_level}. Note that Mixtral and Llava3 graders have disadvantage since they cannot take image input for image-related questions.

\begin{table*}[t]
\tablefontsize
\centering
\renewcommand{\arraystretch}{1.2}
\begin{tabular}{ll|ccc|ccc}
\toprule
        &        & \multicolumn{3}{c}{without ref}                                                                   & \multicolumn{3}{c}{with ref}                                                                      \\
        &        & \multicolumn{1}{l}{same question} & \multicolumn{1}{l}{same exam} & \multicolumn{1}{l}{diff exam} & \multicolumn{1}{l}{same question} & \multicolumn{1}{l}{same exam} & \multicolumn{1}{l}{diff exam} \\ \hline
Mixtral & 0 shot & 0.404                             &                               &                               & 0.445                             &                               &                               \\
        & 1 shot & 0.542                             & 0.549                         & 0.619                         & 0.565                             & 0.390                         & 0.344                         \\
        & 2 shot & 0.564                             & 0.505                         & 0.620                         & 0.500                             & 0.466                         & 0.463                         \\
Llama3  & 0 shot & 0.649                             &                               &                               & 0.677                             &                               &                               \\
        & 1 shot & 0.812                             & 0.731                         & 0.706                         & 0.883                             & 0.770                         & 0.772                         \\
        & 2 shot & 0.771                             & 0.729                         & 0.768                         & 0.788                             & 0.738                         & 0.785                         \\
GPT-4V   & 0 shot & 0.911                             &                               &                               & 0.938                             &                               &                               \\
        & 1 shot & 0.886                             & 0.906                         & 0.893                         & 0.902                             & 0.904                         & 0.948                         \\
        & 2 shot & 0.921                             & 0.910                         & 0.896                         & 0.917                             & 0.888                         & 0.934                        \\
\bottomrule
\end{tabular}
\caption{LLM graders' Pearson correlation to expert graders on exam level, scores normalized. Note that there are only a single scores for zero-shot, since they do not have different shot settings.}
\label{tab:pearson_exam_level}
\end{table*}

\subsection{RMSE on Question-level}
Since the LLM graders are asked to output the scores in the original scale, RMSE would be the most informative metric, since it also reflects the importance of the questions that have higher maximum scores. The LLM graders' performance in RMSE is shown in Table \ref{tab:rmse}.

\label{sec:rmse}
\begin{table*}[t]
\tablefontsize
\centering
\renewcommand{\arraystretch}{1.2}
\begin{tabular}{llcccccc}
\toprule
        &        & \multicolumn{3}{c}{without ref}                                                                   & \multicolumn{3}{c}{with ref}                                                                      \\
        &        & \multicolumn{1}{l}{same question} & \multicolumn{1}{l}{same exam} & \multicolumn{1}{l}{diff exam} & \multicolumn{1}{l}{same question} & \multicolumn{1}{l}{same exam} & \multicolumn{1}{l}{diff exam} \\ \hline
\multicolumn{8}{l}{\textbf{Text-only questions}}                                                                                                                                                                         \\
Mixtral & 0 shot & 3.25                              &                               &                               & 2.96                              &                               &                               \\
        & 1 shot & 2.68                              & 2.90                          & 2.83                          & 2.54                              & 2.65                          & 2.79                          \\
        & 2 shot & 2.69                              & 2.83                          & 2.86                          & 2.51                              & 2.46                          & 2.70                          \\
Llama3  & 0 shot & 2.66                              &                               &                               & 2.09                              &                               &                               \\
        & 1 shot & 1.89                              & 2.30                          & 2.50                          & 1.45                              & 1.92                          & 1.85                          \\
        & 2 shot & 1.88                              & 2.36                          & 2.31                          & 1.92                              & 1.77                          & 1.92                          \\
GPT-4V   & 0 shot & 1.56                              &                               &                               & 1.20                              &                               &                               \\
        & 1 shot & 1.34                              & 1.32                          & 1.53                          & 1.49                              & 1.34                          & 1.28                          \\
        & 2 shot & 1.25                              & 1.29                          & 1.54                          & 1.36                              & 1.31                          & 1.31                          \\ \hline
\multicolumn{8}{l}{\textbf{Image-related questions}}                                                                                                                                                                     \\
GPT-4V   & 0 shot & 0.30                              &                               &                               & 0.28                              &                               &                               \\
        & 1 shot & 0.22                              & 0.24                          & 0.29                          & 0.19                              & 0.26                          & 0.17                          \\
        & 2 shot & 0.23                              & 0.28                          & 0.30                          & 0.18                              & 0.24                          & 0.19                         \\
\bottomrule
\end{tabular}

\caption{LLM grading's RMSE compared to expert grading on the question level. Note that there are only single scores for zero-shot, since they do not have different shot settings.}
\label{tab:rmse}
\end{table*}

\subsection{Performance on Giving Full Points} \label{sec:grader_full}
The performance of the LLM graders on assigning full points to the answers is shown in Table \ref{tab:Mixtral_max_score}.

\begin{table}[htbp]
\tablefontsize
\centering
\begin{tabular}{llrrcr}
\toprule
        &   & \multicolumn{2}{c}{\begin{tabular}[c]{@{}c@{}}Max score\\ precision\end{tabular}} & \multicolumn{2}{c}{\begin{tabular}[c]{@{}c@{}}Max score\\ predicted (\%)\end{tabular}} \\ \cline{3-6} 
        &   & \multicolumn{1}{c}{No ref}                & \multicolumn{1}{c}{Ref}               & No ref                            & \multicolumn{1}{c}{Ref}                            \\
Mixtral & 0 & 0.181                                     & 0.196                                 & 67.0                              & 37.2                                               \\
        & 1 & 0.251                                     & 0.325                                 & 46.3                              & 26.6                                               \\
        & 2 & 0.258                                     & 0.256                                 & 45.3                              & 37.1                                               \\
Llama3  & 0 & 0.380                                     & 0.636                                 & 19.1                              & 5.9                                                \\
        & 1 & 0.438                                     & 0.579                                 & 19.9                              & 8.5                                                \\
        & 2 & 0.447                                     & 0.483                                 & 17.8                              & 10.5                                               \\
GPT-4V   & 0 & 0.527                                     & 0.405                                 & 15.0                              & 11.3                                               \\
        & 1 & 0.422                                     & 0.526                                 & 16.8                              & 10.2                                               \\
        & 2 & 0.458                                     & 0.560                                 & 15.9                              & 9.8                                               \\
\bottomrule
\end{tabular}
\caption{Performance on giving full points}
\label{tab:Mixtral_max_score}
\end{table}

\end{document}